\documentclass[letterpaper, 10 pt, conference]{ieeeconf}  

\IEEEoverridecommandlockouts                              

\usepackage{amsmath,amsfonts}
\usepackage{algorithmic}
\usepackage{array}
\usepackage{arydshln}
\usepackage{textcomp}
\usepackage{stfloats}
\usepackage{float}
\usepackage{url}
\usepackage{verbatim}
\usepackage{graphicx}
\def\BibTeX{{\rm B\kern-.05em{\sc i\kern-.025em b}\kern-.08em
    T\kern-.1667em\lower.7ex\hbox{E}\kern-.125emX}}
\usepackage{balance}
\usepackage{graphics} 
\usepackage{wrapfig}
\usepackage{stackengine}
\usepackage{subcaption}
\usepackage{mwe}
\usepackage[british]{babel}
\usepackage{hhline}
\usepackage{multirow}
\usepackage{amssymb}
\usepackage{xcolor,pifont}
\usepackage{tikz}

\usepackage[version=4]{mhchem}

\usepackage{hyperref}
\usepackage{caption}
\usepackage{comment}
\usepackage{booktabs}

\newcommand*\colourcheck[1]{%
  \expandafter\newcommand\csname #1check\endcsname{\textcolor{#1}{\ding{52}}}%
  }
\colourcheck{blue}
\colourcheck{green}
\colourcheck{red}
\definecolor{mypink1}{rgb}{0.858, 0.188, 0.478}
\definecolor{mypink2}{RGB}{219, 48, 122}
\definecolor{mypink3}{cmyk}{0, 0.7808, 0.4429, 0.1412}
\definecolor{red}{RGB}{255, 0, 0}

\definecolor{blackk}{RGB}{0,0,0}
\definecolor{orangee}{RGB}{230,159,0}
\definecolor{skybluee}{RGB}{86, 180, 233}
\definecolor{bluishgreenn}{RGB}{0,158,115}
\definecolor{yelloww}{RGB}{240,228,66}
\definecolor{bluee}{RGB}{0,114,178}
\definecolor{vermillionn}{RGB}{213,94,0}
\definecolor{reddishpurplee}{RGB}{204,121,167}

\colourcheck{bluishgreenn}

\title{\LARGE \bf
REFNet++: Multi-Task Efficient Fusion of Camera and Radar \\Sensor Data in Bird's-Eye Polar View}

\author{Kavin Chandrasekaran$^{1,2}$, Sorin Grigorescu$^{1,3}$, Gijs Dubbelman$^{2}$, Pavol Jancura$^{2}$
\thanks{$^{1}$Elektrobit Automotive GmbH 
            {\tt\small{\{kavin.chandrasekaran, 
            sorin.grigorescu\}@elektrobit.com}}}
\thanks{$^{2}$Eindhoven University of Technology
            {\tt\small{\{k.chandrasekaran, g.dubbelman, p.jancura\}@tue.nl}}}
\thanks{$^{3}$Transilvania University of Brasov
            {\tt\small{\{s.grigorescu\}@unitbv.ro}}}
}

\begin{document}

\maketitle
\thispagestyle{empty}
\pagestyle{empty}

\begin{abstract}
A realistic view of the vehicle's surroundings is generally offered by camera sensors, which is crucial for environmental perception. Affordable radar sensors, on the other hand, are becoming invaluable due to their robustness in variable weather conditions. However, because of their noisy output and reduced classification capability, they work best when combined with other sensor data. Specifically, we address the challenge of multimodal sensor fusion by aligning radar and camera data in a unified domain, prioritizing not only accuracy, but also computational efficiency. Our work leverages the raw range-Doppler (RD) spectrum from radar and front-view camera images as inputs. To enable effective fusion, we employ a variational encoder-decoder architecture that learns the transformation of front-view camera data into the Bird's-Eye View (BEV) polar domain. Concurrently, a radar encoder-decoder learns to recover the angle information from the RD data that produce Range-Azimuth (RA) features. This alignment ensures that both modalities are represented in a compatible domain, facilitating robust and efficient sensor fusion. We evaluated our fusion strategy for vehicle detection and free space segmentation against state-of-the-art methods using the RADIal dataset.

\end{abstract}

\section{INTRODUCTION}
\label{introductionsec}
Recent Autonomous Driving (AD) technology has been primarily driven by advancements in artificial intelligence, particularly deep learning~\cite{kuutti_survey_2019}. These systems rely on a diverse array of sensors to enable accurate environmental perception. Modern vehicles are built with several sensors, especially cameras, radars, and LiDARs~\cite{sonko_comprehensive_2024}. 

Usually in clear weather conditions, cameras offer detailed visual information, making them significant for understanding complex scenes. However, in AD technology, it is also important to predict the distances of the objects ahead, where radars are preferred due to their depth perception potential. Alternatively, LiDARs generate dense point cloud data with better resolution in a three-dimensional space. Despite these advantages, both cameras and expensive LiDARs face significant performance degradation in unfavorable environments such as snowstorms, dust, and rain with turbulence~\cite{avidan_radatron_2022}, while also providing low-quality velocity information.

Nevertheless, the point-cloud data processed from the radar is generally sparse and suffers from poor angular resolution. Hence, utilizing raw radar signals presents itself as a practical solution. In particular, when recorded synchronously, camera and radar sensor data exhibit complementary strengths, making their fusion an attractive approach to address perception challenges, specifically in segmentation and object detection. However, effectively leveraging raw radar sensor information, especially for fusion with other sensor modalities, continues to be a significant hurdle.

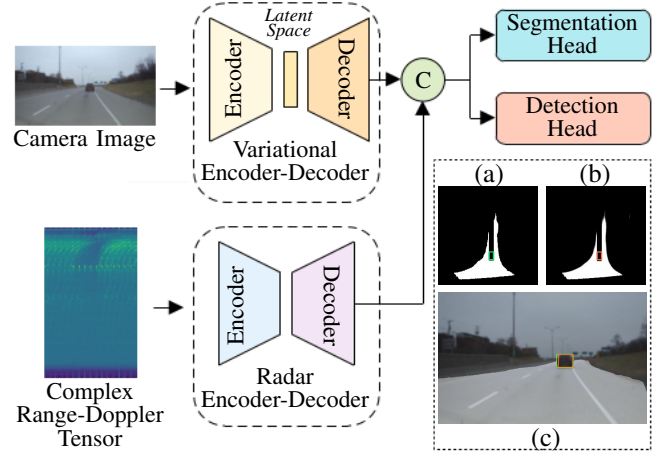
\begin{figure}
    \centering
    \begin{tikzpicture}
        \node[anchor=north west] (base) at (0, 0) 
            {\includegraphics[width=8.5cm, height=6.04cm]  {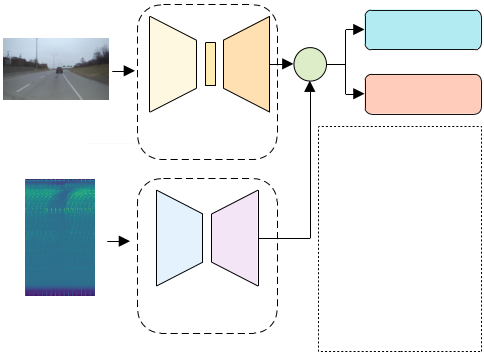}};
    
        \node[anchor=north west] at (5.66, -2.5) 
            {\includegraphics[width=1.3cm, height=1.3cm]{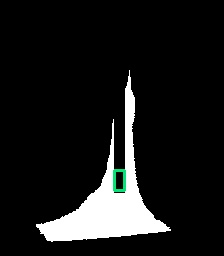}};
        \node[anchor=north west] at (7.09, -2.5) 
            {\includegraphics[width=1.3cm, height=1.3cm]{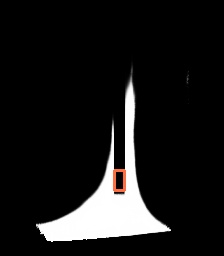}};
        \node[anchor=north west] at (5.66, -3.9) 
            {\includegraphics[width=2.74cm, height=1.74cm]{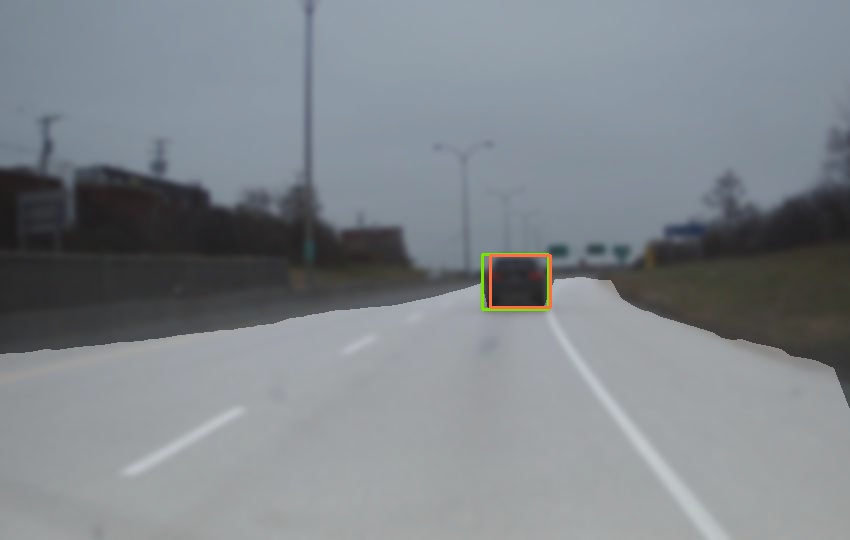}};

        \node[anchor=north] at (1.1, -1.75) {\small{Camera Image}};
        \node[anchor=north] at (1.17, -5.1) {\small{Complex}};
        \node[anchor=north] at (1.17, -5.4) {\small{Range-Doppler}};
        \node[anchor=north] at (1.17, -5.72) {\small{Tensor}};
        \node[anchor=north] at (3.8, -0.15) {\scriptsize\textit{Latent}};
        \node[anchor=north] at (3.8, -0.35) {\scriptsize\textit{Space}};
        \node[anchor=north] at (3.75, -1.9) {\small{Variational}};
        \node[anchor=north] at (3.75, -2.2) {\small{Encoder-Decoder}};
        \node[anchor=north] at (3.75, -4.9) {\small{Radar}};
        \node[anchor=north] at (3.75, -5.2) {\small{Encoder-Decoder}};
        \node[anchor=north] at (7.55, -0.22) {\small{Segmentation}};
        \node[anchor=north] at (7.55, -0.52) {\small{Head}};
        \node[anchor=north] at (7.55, -1.32) {\small{Detection}};
        \node[anchor=north] at (7.55, -1.62) {\small{Head}};
        \node[anchor=north, rotate=90] at (2.83, -1.2) {\small{Encoder}};
        \node[anchor=north, rotate=90] at (2.86, -4.2) {\small{Encoder}};
        \node[anchor=north, rotate=270] at (4.8, -1.2) {\small{Decoder}};
        \node[anchor=north, rotate=270] at (4.6, -4.2) {\small{Decoder}};
        \node[anchor=north] at (5.58, -0.98) {\small{C}};
        \node[anchor=north] at (6.45, -2.2) {(a)};
        \node[anchor=north] at (7.85, -2.2) {(b)};
        \node[anchor=north] at (7.2, -5.7) {(c)};
        
    \end{tikzpicture}
    \caption{\textbf{Architectural Overview:} The variational encoder-decoder architecture \textit{learns} the transformation from the front-view camera image to BEV, which corresponds to the RA domain. On the other hand, the radar encoder-decoder architecture \textit{learns} to recover the angle information from the complex range-Doppler (RD) input, producing RA features. Free space segmentation and vehicle detection are performed on the resulting features fused by concatenation by the appropriate heads. (a) ground truth labels, (b) prediction results, and (c) predictions projected onto the camera image.}
    \label{figarch:introoverview}
\end{figure}

In recent times, BEV-based methods~\cite{wang_c4rfnet_2025, yang_bevformer_2023} have gained a lot of traction in computer vision tasks. An approach~\cite{lim_radar_2019} is to transform the camera data to BEV and then fuse them with radar RA tensors. Alternatively, both image and radar point cloud data are projected to the BEV space in~\cite{zhou_bridging_2023}, where shared features are learned through independent feature extractors. EA-LSS~\cite{hu_ea-lss_2023} project two-dimensional image features into the 3D BEV coordinates, while~\cite{zhou_cross-view_2022} iteratively extract features from multi-view images through BEV queries. Building on these systems, BEVFusion~\cite{liu_bevfusion_2023} highlights the advantages of transforming features to the BEV domain for multi-sensor fusion, achieving outstanding results. Despite these advancements, there remains significant interest among researchers in using raw sensor information for perception tasks. In this paper, we fuse the raw radar data with camera images from the RADIal dataset~\cite{rebut_raw_2022} as shown in Fig.~\ref{figarch:introoverview}.


Our contributions are the following: 
\begin{itemize}
  \item Our proposed fusion architecture, REFNet++, performs object detection and free space segmentation simultaneously, while our previous work, Resource Efficient Fusion Network, REFNet~\cite{chandrasekaran_resource_2024} focuses only on detection.
  \item We are implicitly learning the transformation using a variational encoder-decoder approach that encodes the front-view camera data and eventually decodes it into a two-dimensional Bird's-Eye polar view, while REFNet performs this image transformation manually.
  \item Our analysis demonstrates that we not only outperform the free space segmentation task and closely compete in detection but are also computationally efficient. 
  \item Moreover, our \textit{upgraded} multi-tasking architecture trains about twice as fast as the REFNet model. The code is made publicly available via \href{https://github.com/tue-mps/refnetplusplus}{https://github.com/tue-mps/refnetplusplus}.
\end{itemize}






\section{RELATED WORK} \label{relatedwork}
\subsection{Camera-Radar dataset} \label{relatedwork_dataset}

High-quality, large-scale, temporally, and spatially synchronized fusion datasets are essential for training deep learning algorithms. There has recently been an increase in interest in using raw radar data~\cite{giroux_t-fftradnet_2023, yang_adcnet_2023, jin_cross-modal_2023} for fusion with camera images to enhance segmentation~\cite{wu_sparseradnet_2024} and detection capabilities~\cite{liu_rofusion_2023, liu_echoes_2023}. Consequently, some efforts were made to provide radar data in the form of Analog Digital Converter (ADC) signal~\cite{roldan_deep_2024, lim_radical_2021}, Range-Azimuth Doppler (RAD) cube~\cite{ouaknine_carrada_2021, zhang_raddet_2021}, Range-Azimuth (RA) maps~\cite{sheeny_radiate_2021}, or range-Doppler (RD) spectrum~\cite{mostajabi_high_2020}. Further detailed comparisons among such datasets can be referred to~\cite{yao_radar-camera_2023, wang_multi-sensor_2020}.

Our motivation is to perform both free space segmentation and object detection using raw radar information due to its overarching representation. Hence, the relevant datasets that we examined are RADIal~\cite{rebut_raw_2022}, Radatron~\cite{avidan_radatron_2022}, RADDet~\cite{zhang_raddet_2021}, CARRADA~\cite{ouaknine_carrada_2021}, K-Radar~\cite{paek_k-radar_2023} and RaDelft~\cite{roldan_deep_2024}. The recent K-Radar and RaDelft dataset contains radar data in 4D in various challenging weather conditions. However, the effect of such scenarios is not the focus in this work.

RADIal~\cite{rebut_raw_2022} has been selected for this research due to its unique nature consisting of all radar forms, starting from ADC signals to point cloud data, and well synchronized with camera, LiDAR and odometry. This indicates that there is a substantial possibility of investigating different fusion strategies using this dataset. Although only vehicles can be detected, our work can be further extended to include other classes, provided that an appropriate dataset. 



\subsection{Object detection via camera-radar fusion} \label{relatedwork_crmethods_det}

Radar point cloud methods face challenges due to low angular resolution and sparsity. However, the use of RAD tensors raises issues with computation and storage. Thus, RTCNet~\cite{palffy_cnn_2020} divides the RAD tensors into tiny cubes to minimize computational effort with the use of 3D CNNs. Besides that, in order to extract spatial information, the networks of~\cite{rebut_raw_2022, zhang_object_2020} take in the complex range-Doppler (RD) spectrum. By employing RA maps for detection, RODNet~\cite{wang_rodnet_2021} prevents false alarms caused by Doppler profiles. Other recent methods~\cite{roldan_deep_2024, paek_k-radar_2023} include elevation information contained in a 4D cube that requires a huge pre-processing budget. With little success, the use of ADC data has recently drawn attention~\cite{giroux_t-fftradnet_2023, yang_adcnet_2023}.


The previously stated techniques use raw radar data as a standalone input. However, a recent survey~\cite{shi_radar_2024} summarized numerous other camera-radar architectures for detection by comparing fusion techniques at different stages such as early~\cite{lei_hvdetfusion_2023}, late~\cite{long_radiant_2023} or multi-level fusion~\cite{kim_rcm-fusion_2024}. But the models that exploit raw radar information with camera images are our focus. To this end, ROFusion~\cite{liu_rofusion_2023}, Cross Modal Supervision (CMS)~\cite{jin_cross-modal_2023}, EchoFusion~\cite{liu_echoes_2023}, and REFNet~\cite{chandrasekaran_resource_2024} are recent architectures that belong to this scope. 

ROFusion~\cite{liu_rofusion_2023} employs a point-wise fusion strategy that combines image and RD features by leveraging the radar points present inside the bounding box labels. CMS~\cite{jin_cross-modal_2023} takes the RD spectrum as input and relies on pseudo-labels created from camera data. EchoFusion, on the other hand, fuses Range-Time (RT) radar data and camera images in BEV using the polar-aligned attention technique~\cite{liu_echoes_2023}. Finally, our previous architecture, REFNet~\cite{chandrasekaran_resource_2024} proposes an independent image processing pipeline to transform camera data to the BEV Polar domain. Then it is fused with the RA features obtained from the radar decoder. In contrast, we do not perform any separate pre-processing to save intended costs. Instead, we affirm that our \textit{camera only} architecture implicitly learns this transformation. 


Section~\ref{method} elucidates the details of the proposed fusion architecture, while Section~\ref{experiments} explains the experimental setup. Previous research often lacked a comprehensive study on the resource requirements of their models. Hence, we compare our method on both the accuracy and the computational efficiency as tabulated in Section~\ref{results}.



\subsection{Free space segmentation via camera-radar fusion}\label{relatedwork_crmethods_seg}
A taxonomy of algorithms~\cite{rebut_raw_2022, giroux_t-fftradnet_2023, yang_adcnet_2023, jin_cross-modal_2023, wu_sparseradnet_2024} performs free space segmentation and detection in multitasking mode. Methods such as Occupany grid map learning~\cite{jin_semantic_2024} and TransRadar~\cite{dalbah_transradar_2024} focus mainly on free space segmentation using only the raw range-Doppler (RD) data. Quite surprisingly, there exists no camera and raw radar fusion method specifically designed for free space segmentation using RADIal dataset due to its complex nature. 




\section{METHOD} \label{method}

\subsection{Problem statement} \label{problemstatement}
The problem statement focuses on developing an architecture that efficiently addresses the proposed perception tasks by presenting a resource-conscious fusion strategy.

\subsection{Background} \label{imageprocessing}

The ADC signals from the radar can be processed to obtain various other forms of representation, as stated in Section~\ref{relatedwork_dataset} and detailed in~\cite{yao_exploring_2024}. Alternatively, the camera sensor typically records the data in perspective view. Finding a common representation is therefore essential to achieve sensor fusion. Known for its complexity and higher resolution, processing raw radar data demands a large amount of computing power. However, the height at which the camera sensor is placed and the pitch value play a crucial role while transforming the recorded camera images into a BEV object. In addition, approximation in intrinsic and extrinsic calibration parameters of the camera and imprecise synchronization between radar and camera could affect the transformation. In order to bypass such downsides, we propose a variational encoder-decoder architecture that learns the transformation of front-view camera data into the BEV polar domain while the radar architecture learns to recover the angle information from the RD inputs producing RA features.

\begin{figure*}[!ht]
    \centering 
    \includegraphics[width=\textwidth]{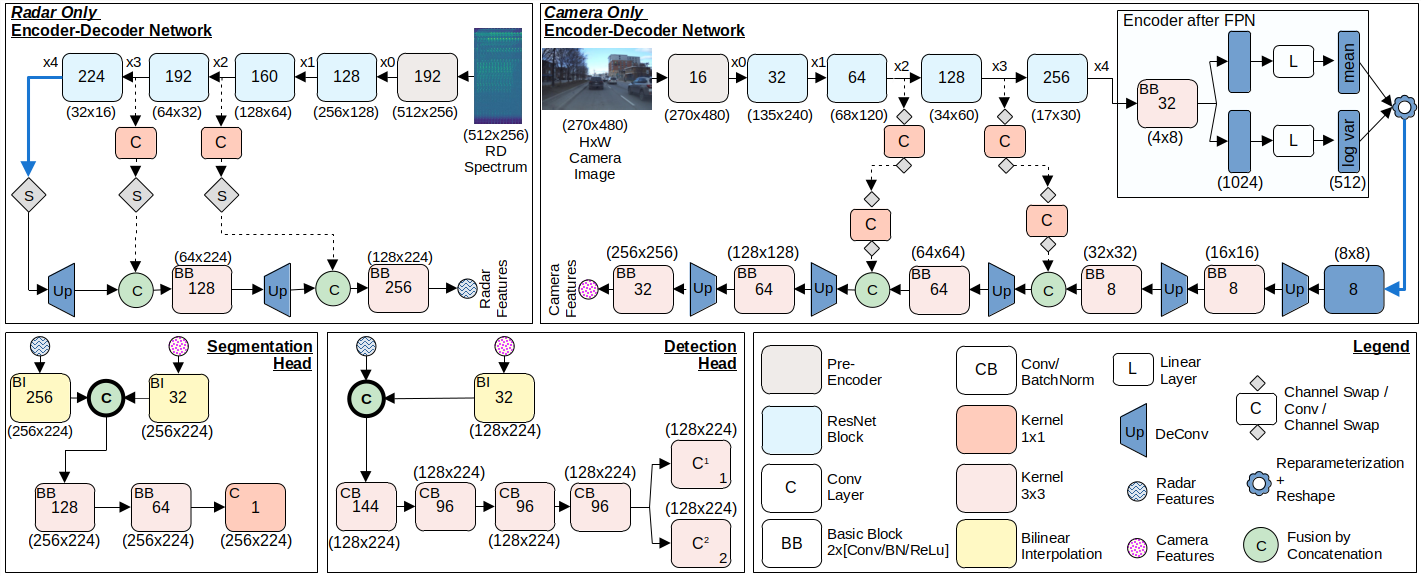}
    \caption{\textbf{Multitask resource-efficient fusion architecture:} The input to the \textit{radar only} network is the range-Doppler (RD) data, while the \textit{camera only} network intakes front-view camera images. x0 to x4 are the feature maps from the respective encoder blocks. The encoder is connected to the decoder by a thick blue arrow, by which the encoded features are upscaled to higher resolutions. The skip connections are shown as dotted lines that preserves the spatial information. The radar and camera features are fused by concatenation on the subsequent heads. Predictions are in Bird's Eye RA Polar View as shown in Fig.~\ref{figarch:introoverview} and~\ref{fig:qualitativeresults_fusion}.}
    \label{figarch:afterdecoderfusion}
\end{figure*}


\subsection{Architecture design} \label{refnetplusplus}

Fig.~\ref{figarch:afterdecoderfusion} shows our deep fusion architecture, detailing the radar \textit{(top-left block)} and camera \textit{(top-right block)} networks.

\textbf{\textit{Radar network:}} Processing the raw ADC signal or range-azimuth Doppler (RAD) 3D tensor has a higher computational overhead. Using a denser range-Doppler (RD) map is another option to take into account, particularly if angle information can be recovered from RD maps~\cite{rebut_raw_2022}.

There are 16 channels in the input RD tensor, since the radar sensor employed has 16 receiving antennas. This implies that each receiving antenna perceives 12 times the signature of any object, such as a car in front, since there are 12 transmitting antennas in the setup. It will be measured specifically at range-Doppler positions $\{R, (D+k\Delta)[D_{\text{max}}]\}_{k=1}^{k=12}$, where $\Delta$ represents the Doppler shift brought on by the transmitted signal's phase shift $\Delta_\varphi$. $D_{max}$ is the maximum measurable Doppler value. All measured Doppler values $(D+k\Delta)$ must fall within this range. If the measured Doppler value is greater than $D_{\text{max}}$, it will be trimmed to fit within $D_{\text{max}}$.

As a result, the Range and Doppler values are represented as complex numbers $(R+iD)$ in the range-Doppler (RD) input tensor. The input contains 32 channels with 512 and 256 range and Doppler bins, respectively, when the real and imaginary components of this tensor are rearranged and concatenated. In order to achieve this, we adapt a Multi-Input Multi-Output (MIMO) pre-encoder~\cite{donnet_mimo_2006} that restructures this RD input into a comprehensible representation for encoder blocks that have 3, 6, 6, and 3 residual layers, respectively, drawing inspiration from FFTRadNet~\cite{rebut_raw_2022}. 

In particular, the encoded feature maps can be seen as azimuth, range, and Doppler, respectively. Prior to upscaling the feature maps, the Doppler and azimuth axes are swapped using the \textit{channel swapping strategy}, as the objective is to obtain angle information. This is illustrated with a rhombus shape in Fig.~\ref{figarch:afterdecoderfusion}. Thus, we aim to learn dense RA feature maps, highlighting their significance for downstream detection and segmentation tasks.

\textbf{\textit{Camera network:}} The camera feature extractor uses a variational encoder-decoder architecture~\cite{lu_monocular_2019} that learns the transformation of front-view camera images to Bird's-Eye view. It is important to note that the dimensions of the camera image input are reduced to ${(3{\times} 270{\times} 480)}$, which is one-fourth of the original image size from the dataset to save memory and processing time.


The pre-encoder block of the \textit{camera only} architecture begins by performing an initial feature extraction using a regular kernel size of 3. The following four blocks that comprise our Feature Pyramid Network (FPN) encoder also include 3, 6, 6, and 3 residual layers, respectively. Each encoder block executes a ${2{\times} 2}$ downsampling that leads to a reduction in tensor size by a factor of about 16 in width and height. The downsampling is done to systematically reduce the spatial resolution so that the encoder captures essential features efficiently.


The encoder after FPN processes the resulting feature maps using convolutional and linear layers that return the mean ($\mu$) and log variance ($\log[\sigma^2]$). $\mu$ is the central point in the latent space, and $\log[\sigma^2]$ is the uncertainty or spread of the latent distribution. This is encoded as a 512-dimensional vector. Using the reparameterization trick~\cite{lu_monocular_2019, qin_epanechnikov_2024}, the model samples the latent vector as:
\begin{align}
z = \mu + \sigma \cdot \epsilon, \quad \epsilon \sim \mathcal{N}(0, 1),
\end{align}where, $\sigma = \exp\left(0.5 \cdot \log\left[\sigma^2\right]\right)$ is the standard deviation derived from the logarithm of variance; $\epsilon \sim \mathcal{N}(0, 1)$ is a random variable chosen from a standard normal distribution with mean 0 and variance 1. More specifically, this introduces randomness ($\epsilon$) to encourage the model to explore various latent representations around $\mu$. This stochasticity helps the network to generalize better and avoid over-fitting by modeling the inherent variability in the data.

Moreover, instead of directly sampling ($z \sim \mathcal{N}(\mu, \sigma^2)$), which is non-differentiable because of the stochastic nature of sampling, we use $z = \mu + \sigma \cdot \epsilon$, which allows the gradients to flow through $\mu$ and $\sigma$ during backpropagation. During inference, the function bypasses the sampling step and simply returns the mean ($z = \mu$), since during inference, the network should behave deterministically, using the most probable value (the mean) of the latent variable.

To reconstruct the desired BEV representation, the decoder then takes the latent vector $z$ and the FPN feature maps through skip connections as shown by the dotted lines in Fig.~\ref{figarch:afterdecoderfusion}. It can be noticed that there are four convolutional layers: two for processing the $\text{x2}$ feature maps and another two dedicated to the $\text{x3}$ feature maps. There is also a channel swap symbol placed just before and after the convolutional layers. This channel swapping is different from what has been done in the \textit{radar only} part.

Fusion by concatenation along the channel axis is feasible only when the feature dimensions are aligned. Hence, the $\text{x3}$ features are first swapped ${(128{\times} 34{\times} 60 \stackrel{\text{swap}}{\longrightarrow} 60{\times} 34{\times} 128)}$ so that the convolutional layer appropriately transforms the dimensions ${(60{\times} 34{\times} 128 \stackrel{\text{conv2d}}{\longrightarrow} 32{\times} 34{\times} 128)}$. Further, it is again swapped back so that the camera image feature is retained ${(32{\times} 34{\times} 128 \stackrel{\text{swap}}{\longrightarrow} 128{\times} 34{\times} 32)}$. In order to match the dimensions of the decoder features, there is a need to repeat the process again. This strategy allows one to view the encoder feature map in a dimension ${(128{\times} 34{\times} 32 \stackrel{\text{swap/conv2d/swap}}{\longrightarrow} 128{\times} 32{\times} 32)}$ that aligns with the decoder features for concatenation. The same process is applied to the $\text{x2}$ feature maps. This preparation helps the network in effectively fusing with the radar features within the segmentation and detection head during training, thus reducing computational overhead. The network's backbone can be replaced by heavier models based on resource availability.

\textbf{\textit{Detection head:}} The bilinear interpolation technique~\cite{noauthor_torchnnfunctionalinterpolate_nodate} is used to align the dimensionality of camera-decoded features. The RA latent features from the \textit{radar only} network is then fused with the camera features by channel concatenation. Further processing is carried out using four Convolution-BatchNorm layers with 144, 96, 96 and 96 filters, respectively. 

The branch finally splits for classification and regression. A ${3{\times} 3}$ convolutional layer with sigmoid activation makes up the classification part (C$^{1}$), which predicts a probability map. Every pixel on this map represents a binary classification of whether or not a vehicle is present. The predictions from the classification part have dimension ${1{\times} 128{\times} 224}$, with a resolution of 0.8m in range and 0.8$^{\circ}$ in azimuth. The same ${3{\times} 3}$ convolution layer is used in the regression part (C$^{2}$), which produces two feature maps. One channel denotes the range, while the other denotes the azimuth predictions of the objects detected. 



We apply the focal loss~\cite{lin_focal_2018} to the classification output, since a significant amount of the scene is background. This stabilizes the training process and prevents the problem of class imbalance. For positive detections, the regression output is subjected to smooth L1 loss. 
\begin{equation}
\mathcal{L}_{det} = \text{Focal}(y_{\text{cls}}, \hat{y}_{\text{cls}}) + \alpha \text{Smooth-L1}(y_{\text{reg}}, \hat{y}_{\text{reg}}),
\end{equation} where $y_{\text{cls}}$ and $y_{\text{reg}}$ are the ground truths for the classification and regression parts, respectively. $\hat{y}_{\text{cls}}$ and $\hat{y}_{\text{reg}}$ represent the predicted values. $\alpha$ is a positive hyperparameter that symmetrize the contributions of the two loss functions. 

\textbf{\textit{Segmentation head:}} Using bilinear interpolation, the radar- and camera-decoded feature maps have their dimensions aligned. The fusion is performed through channel concatenation, which is represented by a thick black green circle in Fig.~\ref{figarch:afterdecoderfusion}. The resulting RA maps are processed by two basic blocks. Each of these blocks performs the Convolution-BatchNorm-ReLu operation twice. Finally, a 1x1 convolution produces a 2D output feature map with the probability that each pixel location is drivable. The dimensions of the prediction are ${1{\times} 256{\times} 224}$ with a resolution of 0.4m in the range which corresponds to half of the original range and 0.2$^{\circ}$ in azimuth (within [$-45^\circ$, $45^\circ$], which is only \text{50\%} of the total azimuth field-of-view). The binary cross entropy loss (BCE) is applied to the free space segmentation head: 
\begin{equation}
\mathcal{L}_{\text{seg}} = 
\sum_{(r, a) \in \Psi} \text{BCE}(y_{\text{seg}}(r, a), \hat{y}_{\text{seg}}(r, a)),
\end{equation} where $\Psi = [1, B_R/2] \times [1, B_A/4]$, $y_{\text{seg}}$ represents the one-hot ground truth and $\hat{y}_{\text{seg}}$ are the prediction maps. $B_R$ and $B_A$ are range and azimuth bins respectively.

\textbf{\textit{Combined loss}}: Our architecture can be deployed in a single- or multi-task mode. If both heads are activated, the total Multi-Task Loss (MTL) is the sum of the detection and segmentation losses for every training sample $\mathbf{x}$. 
\begin{equation}
\mathcal{L}_{\text{MTL}} = \sum_{\mathbf{x}} 
\mathcal{L}_{\text{det}}(\mathbf{x}) + 
\beta \mathcal{L}_{\text{seg}}(\mathbf{x}),
\end{equation} where $\beta$ is a positive hyperparameter set empirically that balances the two tasks.

\section{EXPERIMENTAL SETUP} \label{experiments}

\subsection{Dataset} \label{radialdataset}
RADIal~\cite{rebut_raw_2022} is composed of synchronized camera, radar, and LiDAR sensor data that includes GPS information. After roughly 25,000 frames of synchronization between the three sensors, 8,252 frames of them or 9,550 vehicles in total are labeled.

\subsection{Training setup} \label{traindetails}
Nvidia RTX A6000 GPU has been used to train the network. A random split of 70\% of the data is allocated for training and 15\% each is preserved for the validation and the test set. Our network is trained using the multitask loss described in Section~\ref{refnetplusplus}, with $\alpha$ and $\beta$ set to 100 as originally proposed~\cite{rebut_raw_2022}. Training is performed using the Adam optimizer~\cite{kingma_adam_2017} with a batch size of 4, for 100 epochs. The initial learning rate is set to $1e^{-4}$ with a decay of 0.9 for every ten epochs.

\subsection{Evaluation metric} \label{evalmetrics}
\textbf{\textit{Performance:}} Average Precision (AP) and Average Recall (AR), considering an Intersection-over-Union (IoU) threshold of 50\%, are used for object detection. The F1 score is calculated directly from AP and AR: \( F1 = \frac{2 \times \text{AP} \times \text{AR}}{\text{AP} + \text{AR}} \). The range error (RE) and angle error (AE) are computed as in~\cite{rebut_raw_2022}. The mean IoU (mIoU) metric determines the accuracy of the free space segmentation task. Since the road surface is barely visible after 50 meters, the metric is calculated in a truncated range of [0m - 50m].

\textbf{\textit{Computational cost:}} Table~\ref{table:compeff} shows how we compared the computational efficiency between different methods. \# denotes the total number of model parameters in millions. For each frame in the test set, we compute the Frames-per-second value and average it across frames. ($\sigma$) denotes the standard deviation calculated from the FPS values. Lower $\sigma$ reflects stable performance between frames. Moreover, we compare the model size in MB and the consumption of GPU memory in GB as in our previous work~\cite{chandrasekaran_resource_2024}.

\subsection{Baselines}
\label{baselines}
\textit{\textbf{Vehicle detection:}} As our work leverages raw radar information and that we use RADIal dataset, we consider the networks that have been trained on the same as baselines. As reviewed in Section~\ref{relatedwork}, FFTRadNet~\cite{rebut_raw_2022}, TFFTRadNet~\cite{giroux_t-fftradnet_2023}, ADCNet~\cite{yang_adcnet_2023} are some of the architectures that fit into this category. Because we aimed at concatenating camera and radar information in Bird's-Eye Polar View, the Cross-Modal Supervision (CMS)~\cite{jin_cross-modal_2023}, ROFusion~\cite{liu_rofusion_2023}, EchoFusion~\cite{liu_echoes_2023} and REFNet~\cite{chandrasekaran_resource_2024} are closely related for comparing detection performances. 

\textit{\textbf{Free space segmentation:}}
For free space segmentation, since there is no available fusion method, we reimplemented REFNet~\cite{chandrasekaran_resource_2024} with our segmentation head. Additionally, we compared our results with other relevant methods, also trained using raw RD data from the RADIal dataset, such as SparseRadNet~\cite{wu_sparseradnet_2024}, TransRadar~\cite{dalbah_transradar_2024}, and Occugrid~\cite{lu_monocular_2019}.

\begin{table*}[!ht]
\caption{Detection results on the test split. RD, ADC, RT, PC, C, C\textsuperscript{BEV} correspond to Range-Doppler, Analog-To-Digital Converter signal, Range-Time, Point Cloud, Camera data in Front and Bird's-Eye View respectively. The single modality methods are separated from multimodal fusion methods by a dashed line. Best values are in bold and second best are underlined. The values missing are marked by a "-", either unreported in the respective works or due to code unavailability\dag. \textsuperscript{*}Trained using \textit{camera only} block of the proposed architecture. \textsuperscript{§}Trained without reparameterization.}
\label{table:detacc}
\centering 
\renewcommand{\arraystretch}{1}
\begin{tabular}{llllllllllll}
\specialrule{0.5pt}{0pt}{0pt}
\hline
\textbf{} & \textbf{} & \multicolumn{5}{c}{\multirow{1.35}{*}{\centering{Single-task}}} & \multicolumn{5}{c}{\multirow{1.35}{*}{\centering{Multi-task}}}
\\\cmidrule(lr){3-7} \cmidrule(lr){8-12}
Methods & Modality & AP(\%)$\uparrow$ & AR(\%)$\uparrow$ & F1(\%)$\uparrow$ & RE(m)$\downarrow$ & AE($^{\circ}$)$\downarrow$ & AP(\%)$\uparrow$ & AR(\%)$\uparrow$ & F1(\%)$\uparrow$ & RE(m)$\downarrow$ & AE($^{\circ}$)$\downarrow$ \\\hline \addlinespace[3pt]

FFTRadNet~\cite{rebut_raw_2022}             & RD                     & 93.45        & 83.35        & 88.11             & \underline{0.12}            & 0.15              & 96.05       & 82.18              & 88.57        & \textbf{0.11}        & 0.17                                                   \\
TFFTRadNet~\cite{giroux_t-fftradnet_2023}              & ADC                             & 90.80         & 88.31         & 89.54              & 0.15            & {0.13}              &89.6         & \underline{89.5}               &89.5        & 0.15     &  0.13                                                    \\
ADCNet\dag~\cite{yang_adcnet_2023}            & ADC                                 & -           & -           & -              & -            & -             & 95          &   89             & 91.9            &  0.13           & {0.11}                                                         \\

SparseRad\dag~\cite{wu_sparseradnet_2024}      & RD                & -      & -              &-        & -        & -               & 96.00        &  \textbf{91.78}        & \textbf{93.84}             & 0.13         & \underline{0.10}        
\\
\addlinespace[1pt] 
\hdashline[2pt/3pt] 
\addlinespace[3pt] 

CMS\dag~\cite{jin_cross-modal_2023}            & RD\&C                      & -         & -        & -             & -            & -                 & \textbf{96.9}        & 83.49               & 89.69           & 0.45            & -                                              \\

ROFusion~\cite{liu_rofusion_2023}      & RD\&PC\&C                      & 91.13        & \textbf{95.29}        &{93.16}             & 0.13            & 0.21              & -      & -              &-        & -        & -                                                         \\
EchoFusion~\cite{liu_echoes_2023}      & RT\&C                            & \textbf{96.95}        & \underline{93.43}        & \textbf{95.16}             & \underline{0.12}            & 0.18              & -      & -              &-        & -        & -    \\   

REFNet~\cite{chandrasekaran_resource_2024}      & RD\&C\textsuperscript{BEV}                              & 95.75        & {91.35}        & {93.49}             & \textbf{0.11}          & \textbf{0.09}              & 95.97      & 87.32            &91.44       & \underline{0.12}      & \underline{0.10}         
\\\hline \addlinespace[3pt]

Ours\textsuperscript{*}     & C                & 77.29      & 71.29      & 74.16      & 0.13 & 0.15             & 75.21      & 68.93      & 71.93      & 0.13& 0.16   \\

Ours\textsuperscript{§}     & RD\&C                             & 96.33      & 87.58      & 91.74      & \underline{0.12} & \underline{0.10}      & 95.45& 87.13      & 91.10 & \underline{0.12}      & \textbf{0.09}  \\

Ours      & RD\&C                              & \underline{96.92}       & {90.68}        & \underline{93.70}             & \textbf{0.11}          & \textbf{0.09}              & \underline{96.16}       & 89.43              & \underline{92.67}        & \textbf{0.11}         & \textbf{0.09}       

\\\hline    
\specialrule{0.5pt}{0pt}{0pt}
\end{tabular}
\end{table*}

\section{EXPERIMENTS AND RESULTS} \label{results}
We used the same training setup and parameters as the other models from Section~\ref{traindetails} for a fair comparison. We reimplemented all multitasking models with only the detection head, as shown in Table~\ref{table:detacc}, and with only the segmentation head, as shown in Table~\ref{table:segacc}, considering them as single-tasking models.

\textit{\textbf{Ablations:}} Taking into account the \textit{camera only} block of the proposed architecture from Fig.~\ref{figarch:afterdecoderfusion}, we conducted further investigation. Because the model is exposed only to camera data, it is computationally efficient but the accuracies significantly dropped in all settings. Secondly, the complete fusion architecture is trained without the reparameterization trick. Training the model in this mode took us three times as long with no significant reward.
\subsection{Quantitative evaluation}
\textit{\textbf{Vehicle detection performance:}} Like REFNet~\cite{chandrasekaran_resource_2024}, our method outperforms existing architectures in range and angle error, indicating that the predicted objects are precisely localized in the scene. The significant improvement in AE this time is due to the fact that the fusion mechanism, i.e. alignment of radar and camera features, is well calibrated. The more intricate design of EchoFusion~\cite{liu_echoes_2023}, which has almost 3.8 times as many trainable parameters as our model, is the reason for the second-highest F1 score and the marginal variation in AP and AR.

\textit{\textbf{Free space segmentation performance:}} As highlighted in Table~\ref{table:segacc}, we substantially outperform other frameworks. This is because the data is fed into the network without undergoing any physical transformation, rather our network learns this transformation while training.

\textit{\textbf{Multitasking performance:}} The detection and segmentation results in the multi-task mode are also shown in Tables~\ref{table:detacc} and~\ref{table:segacc} respectively. Our focus is to offer a balance between accuracy and computational efficiency, leading to second-best detection performance in almost every aspect but surpassing in the segmentation task.

\begin{table}[!ht]
\centering
\caption{Free space segmentation results on the test split. Abbreviations and details as in Table~\ref{table:detacc}.}
\label{table:segacc}
\centering 
\renewcommand{\arraystretch}{1}
\begin{tabular}{llllllll}
\specialrule{0.5pt}{0pt}{0pt}
\hline
\textbf{} & \textbf{} & \multicolumn{1}{c}{\multirow{1.35}{*}{\centering{Single-task}}} & \multicolumn{1}{c}{\multirow{1.35}{*}{\centering{Multi-task}}}
\\\cmidrule(lr){3-4}
Methods & Modality  & \multicolumn{2}{c}{\begin{tabular}[c]{@{}c@{}}mIoU \\ (\%)\end{tabular} $\uparrow$} \\\hline \addlinespace[3pt] 

FFTRadNet~\cite{rebut_raw_2022}             & RD                     & 77.8        & 74.0              \\
TFFTRadNet~\cite{giroux_t-fftradnet_2023}             & ADC                     & 79.43       & 80.20              \\
ADCNet\dag~\cite{yang_adcnet_2023}             & ADC                     & -      & 78.59              \\

SparseRad\dag~\cite{wu_sparseradnet_2024}       & RD                     & -       & 78.48                       \\  
TransRadar~\cite{dalbah_transradar_2024}             & RD                     & 81.93      & -     \\

Occugrid\dag~\cite{jin_semantic_2024}             & RD                     & {82.2}        & -      

\\
\hdashline[2pt/3pt]
\addlinespace[3pt] 

CMS\dag~\cite{jin_cross-modal_2023}             & RD\&C                    & -      & {80.4}         \\
REFNet~\cite{chandrasekaran_resource_2024}             & RD\&C\textsuperscript{BEV}                      & {85.16}      & {84.84}          
\\\hline \addlinespace[3pt]

Ours\textsuperscript{*}             & C                     & 80.63      & 79.51      \\
Ours\textsuperscript{§}             & RD\&C                     & \underline{86.57}      & \underline{85.09}      \\

Ours            & RD\&C                     & \textbf{88.13}      & \textbf{87.58}         
\\ \addlinespace[0.2pt] \hline
\specialrule{0.5pt}{0pt}{0pt}
\end{tabular}
\end{table}

\subsection{Computational efficiency}
\textit{\textbf{Vehicle detection performance:}} Since FTTRadNet~\cite{rebut_raw_2022} and TFFTRadNet~\cite{giroux_t-fftradnet_2023} are non-fusion models that solely rely on radar data, they are computationally efficient, sacrificing accuracy. Alternatively, models such as ROFusion~\cite{liu_rofusion_2023} and EchoFusion~\cite{liu_echoes_2023} show an inefficient use of GPU memory. Although REFNet~\cite{chandrasekaran_resource_2024} seems to be computationally efficient, they employ an independent image processing pipeline to process the camera data for which the cost is not considered. On the other hand, our fusion model is proven to be an optimal choice to achieve superior results as shown in Table~\ref{table:compeff}.


\textit{\textbf{Free space segmentation performance:}} Even though the single modality TransRadar~\cite{dalbah_transradar_2024} architecture seem to possess less computational complexity, it is far behind us in terms of mIoU. As discussed, REFNet~\cite{chandrasekaran_resource_2024} does not consider the computational effort involved while pre-processing the image data. We affirm that our approach also offers the best trade-off in the free space segmentation task.

\textit{\textbf{Multitasking performance:}} In multi-task mode, the models are trained with both detection and segmentation heads. It is also essential to compare the computational cost with single-modality models, ensuring that our fusion approach remains competitive in terms of processing speed, stability, and resource consumption.


\begin{table*}[!ht]
\caption{Computational efficiency. Only models for which computational efficiency could be measured are included due to code availability. Abbreviations and details as in Table~\ref{table:detacc}.}
\label{table:compeff}
\centering 
\renewcommand{\arraystretch}{1}
\begin{tabular}{llllllllllllllllll}
\specialrule{0.5pt}{0pt}{0pt}
\hline
\textbf{} & \multicolumn{5}{c}{\multirow{1.35}{*}{\centering{Object Detection}}} & \multicolumn{5}{c}{\multirow{1.35}{*}{\centering{Free Space Segmentation}}} & \multicolumn{5}{c}{\multirow{1.35}{*}{\centering{Multi Task}}}
\\\cmidrule(lr){2-6} \cmidrule(lr){7-11} \cmidrule(lr){12-16} 

Methods & 
\#$\downarrow$ & \begin{tabular}[c]{@{}c@{}}Avg \\ FPS\end{tabular}$\uparrow$ & $\sigma$ $\downarrow$ & \begin{tabular}[c]{@{}c@{}}Model\\size\end{tabular}$\downarrow$ & \begin{tabular}[c]{@{}c@{}}GPU \\cost\end{tabular}$\downarrow$ & 
\#$\downarrow$ & \begin{tabular}[c]{@{}c@{}}Avg \\ FPS\end{tabular}$\uparrow$ & $\sigma$ $\downarrow$ & \begin{tabular}[c]{@{}c@{}}Model\\size\end{tabular}$\downarrow$ & \begin{tabular}[c]{@{}c@{}}GPU \\cost\end{tabular}$\downarrow$ & 
\#$\downarrow$ & \begin{tabular}[c]{@{}c@{}}Avg \\ FPS\end{tabular} $\uparrow$ & $\sigma$ $\downarrow$ & \begin{tabular}[c]{@{}c@{}}Model\\size\end{tabular}$\downarrow$ & \begin{tabular}[c]{@{}c@{}}GPU \\cost\end{tabular}$\downarrow$ \\\hline \addlinespace[3pt]

FFTRadNet                 & \textbf{3.23} &  \underline{68.46}                      & 2.19       & \textbf{39.2}             & \underline{2.01}     & \textbf{3.16} & \underline{71.91}       & 1.78     & \textbf{38.4}         & \underline{2.01}     & \textbf{3.28}        & \underline{67.32}        & {2.19}             & \textbf{39.2}            & \underline{2.01}                                                \\
TFFTRad                                      & 9.08 &  54.37                     & 4.28    & 109.5             & {2.04}  & 9.01 & 56.43      & \textbf{1.13}    & 108.7        & 2.04 & 9.13 &  52.41                     & 4.28   & 109.5         & {2.04}                                                      \\

TransRadar                     & -&- &- &- &-         &   \underline{3.31} & {69.23}   & 2.06   & \underline{38.5}           & {2.02}        & - &  -                     & -   & -          & -            \\\addlinespace[1pt] 
\hdashline[2pt/3pt]
\addlinespace[3pt] 

ROFusion             & \underline{3.33} &  56.11                     & 1.55       & 87.2             & 2.87 & -&- &- &- &-     & - &  -                     & -   & -          & -                                                     \\
EchoFusion                             & 25.61        & 38.32       & 2.94          & 102.5         & 4.47  & -&- &- &- &-        & - &  -                     & -   & -          & -            \\

              
REFNet                               & 6.58        & {58.91}      & \underline{1.28}       & {79.8}        & 2.06    &6.49 & 52.04     & 1.98    & 78.7          & 2.06      & 7.28 &  53.05                   & \textbf{1.04}  & {88.2}      & 2.06
\\\hline \addlinespace[3pt]

Ours\textsuperscript{*}                                & 3.43      & \textbf{74.29} & 3.22      & \underline{41.7} & \textbf{1.82}   &3.42      & \textbf{72.22} & 3.05      & 41.6& \textbf{1.82}      & \underline{3.44}      & \textbf{69.32} & 2.94     & \underline{41.9} & \textbf{1.89}   \\

Ours\textsuperscript{§}                                & 6.17     & 56.83      & 1.66     & {74.8} & 2.08     & 6.09& 56.16     & 1.81& 73.9     & 2.08& 6.76     & 53.31& 2.12     & 81.9 & 2.09       \\

Ours                               & 6.69     & 58.42      & \textbf{1.22}         & 81.1         & 2.06   &6.61& 57.75     & \underline{1.52}    & 80.2         & 2.06      & 7.26 &  {54.56}                     & \underline{1.91}   & {88.3}         & 2.06    \\
\hline
\specialrule{0.5pt}{0pt}{0pt}
\end{tabular}
\end{table*}

\subsection{Qualitative Evaluation}
\begin{figure}[!ht]
    \centering
        \begin{subfigure}{0.14\textwidth}
        \includegraphics[width=2.1cm, height=2.4cm, angle=0]{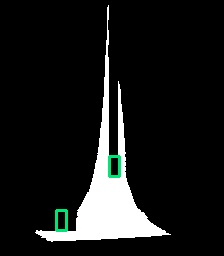}
    \end{subfigure}\hspace{-0.3cm}
    \begin{subfigure}{0.14\textwidth}
        \includegraphics[width=2.1cm, height=2.4cm, angle=0]{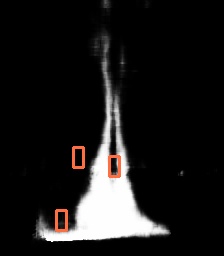}
    \end{subfigure}\hspace{-0.3cm}
    \begin{subfigure}{0.2\textwidth}
        \includegraphics[width=3.78cm, height=2.4cm, angle=0]{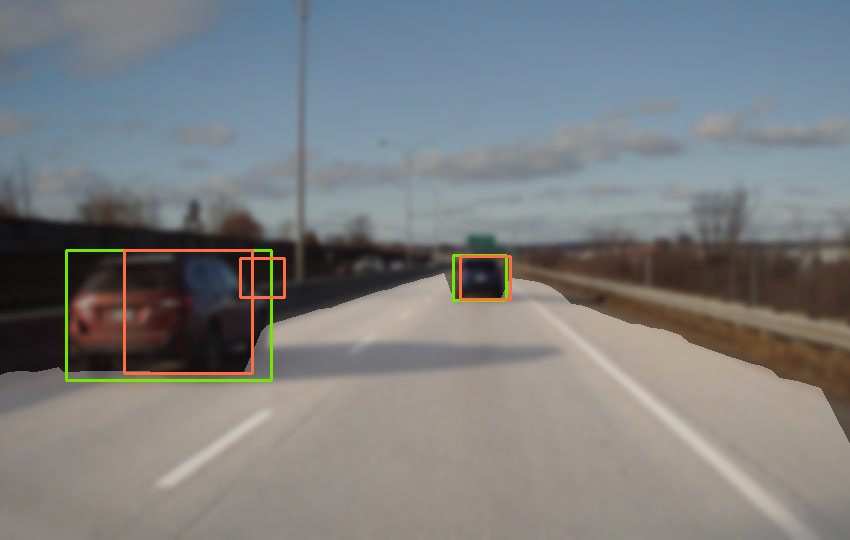}
    \end{subfigure}
    \vspace{0.2cm} 
    
    \begin{subfigure}{0.14\textwidth}
        \includegraphics[width=2.1cm, height=2.4cm, angle=0]{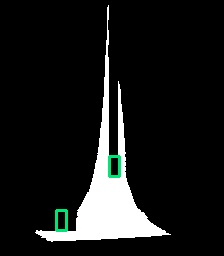}
    \end{subfigure}\hspace{-0.3cm}
    \begin{subfigure}{0.14\textwidth}
        \includegraphics[width=2.1cm, height=2.4cm, angle=0]{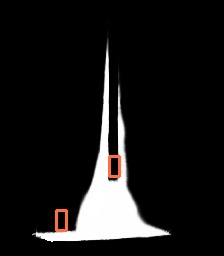}
    \end{subfigure}\hspace{-0.3cm}
    \begin{subfigure}{0.2\textwidth}
        \includegraphics[width=3.78cm, height=2.4cm, angle=0]{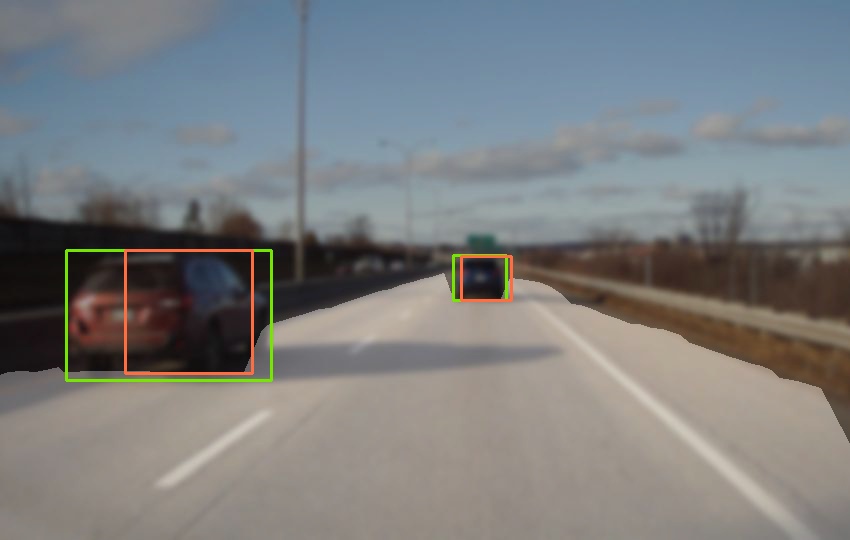}
    \end{subfigure}
    \vspace{0.2cm}
    
    \begin{subfigure}{0.14\textwidth}
        \includegraphics[width=2.1cm, height=2.4cm, angle=0]{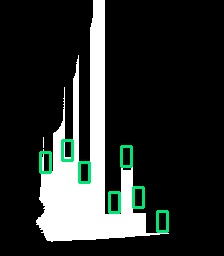}
    \end{subfigure}\hspace{-0.3cm}
    \begin{subfigure}{0.14\textwidth}
        \includegraphics[width=2.1cm, height=2.4cm, angle=0]{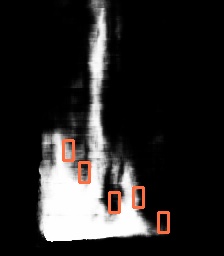}
    \end{subfigure}\hspace{-0.3cm}
    \begin{subfigure}{0.2\textwidth}
        \includegraphics[width=3.78cm, height=2.4cm, angle=0]{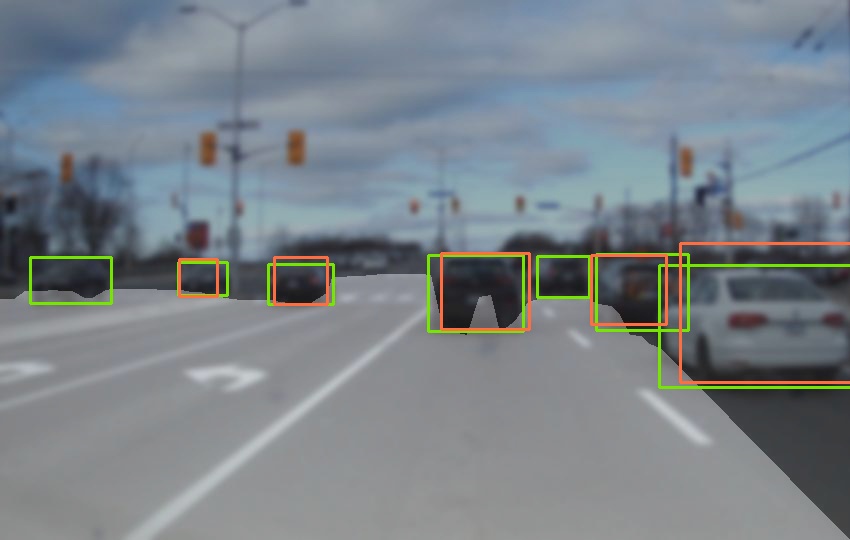}
    \end{subfigure}
    \vspace{0.2cm}
    
    \begin{subfigure}{0.14\textwidth}
        \includegraphics[width=2.1cm, height=2.4cm, angle=0]{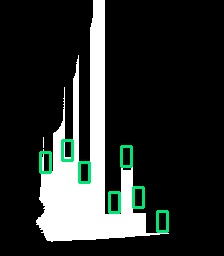}
    \end{subfigure}\hspace{-0.3cm}
    \begin{subfigure}{0.14\textwidth}
        \includegraphics[width=2.1cm, height=2.4cm, angle=0]{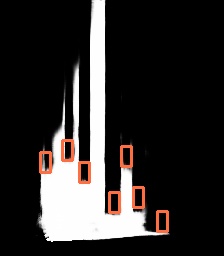}
    \end{subfigure}\hspace{-0.3cm}
    \begin{subfigure}{0.2\textwidth}
        \includegraphics[width=3.78cm, height=2.4cm, angle=0]{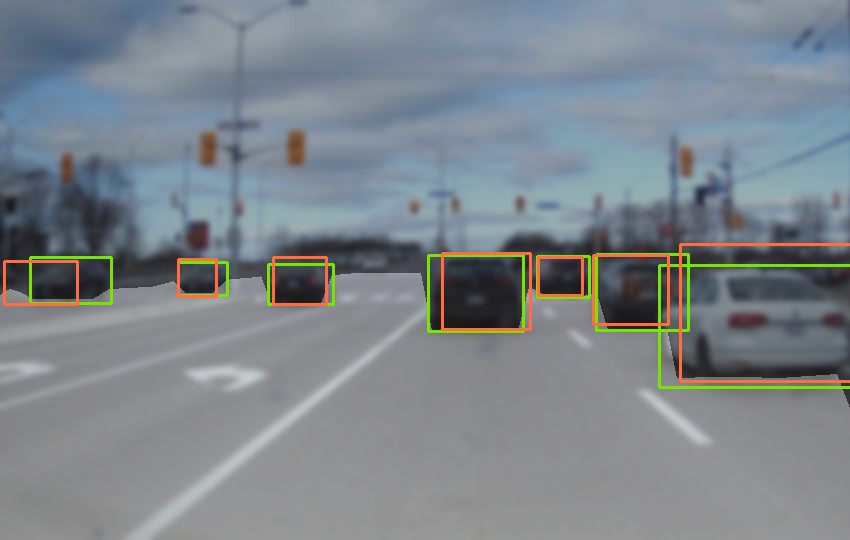}
    \end{subfigure}
    \vspace{0.2cm} 
    \put(-212,-12){\fontsize{10}{10}\selectfont (a)}
    \put(-150,-12){\fontsize{10}{10}\selectfont (b)}
    \put(-52,-12){\fontsize{10}{10}\selectfont (c)}

    \begin{tikzpicture}[overlay, remember picture]
        \node[rotate=90] at (-4.4, 9.5) {\textit{\small Radar only}};
        \node[rotate=90] at (-4.4, 7) {{\small Ours}};
        \node[rotate=90] at (-4.4, 4.35) {\textit{\small Camera only}};
        \node[rotate=90] at (-4.4, 1.7) {{\small Ours}};
    \end{tikzpicture}
    
\caption{Qualitative results on samples from the test set. (a) depicts the ground truth labels in Bird’s-Eye Polar View where the free space segmentation is in white while the BEV bounding boxes are in green. (b) are our prediction results. For a better visualization, the bounding box predictions are projected onto the camera images as shown in (c), with the ground-truth boxes. The free space predictions are projected by consolidating the intensity values. Zoom in to better visualize the False Positives and False Negatives.}
\label{fig:qualitativeresults_fusion}
\end{figure}
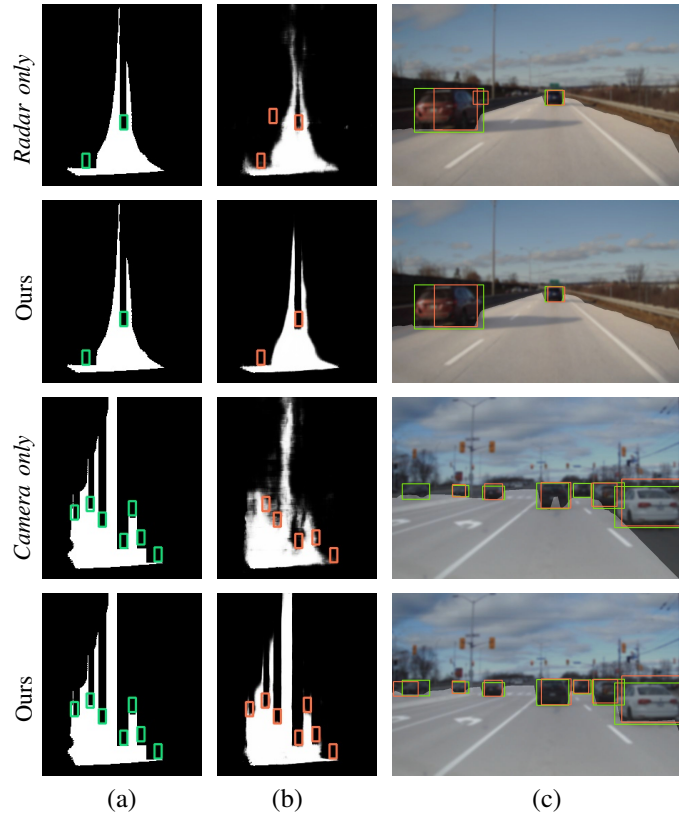

We visualize our fusion results against the results of \textit{radar only} and \textit{camera only} part of our architecture, as illustrated in Fig.~\ref{fig:qualitativeresults_fusion}. The \textit{radar only} model exhibits a false positive detection, whereas the \textit{camera only} model fails due to false negative detections, making it an inadequate solution. Uncertain free space predictions at several pixel locations worsen the situation. These outcomes highlight our fusion model's resilience in producing reliable predictions for both tasks, increasing its applicability in real-world situations.

\section{CONCLUSION AND FUTURE WORK} \label{conclusion}


We proposed REFNet++, a fusion architecture that performs multitasking and can also operate in a single task mode in the BEV domain, designed to boost the computational efficiency of the camera-radar perception system. In line with our research goal and the results demonstrated on the RADIal dataset, our method exhibits excellent trade-off between performance while retaining a comparatively low computing power. 

Furthermore, including other modality like LiDAR remains a viable consideration, contingent upon the assumption that constraints related to cost and memory overhead are deemed non-critical or can be effectively mitigated. However, it takes a lot of effort to obtain large-scale, high-quality, and time-synchronized multimodal data with accurate annotations. Building such a large, diverse dataset is another possible way to speed up future studies.


\bibliographystyle{ieeetr}
\bibliography{references_z.bib}

\end{document}